\pdfoutput=1
\documentclass[11pt]{article}

\usepackage[review]{ACL2023}

\usepackage{graphicx}
\usepackage{times}
\usepackage{latexsym}
\usepackage{multicol}
\usepackage{multirow}
\usepackage{booktabs}
\usepackage{amsmath}
\usepackage{enumitem}
\usepackage{subcaption}

\usepackage[T1]{fontenc}
\usepackage[utf8]{inputenc}

\usepackage{microtype}

\usepackage{inconsolata}
\usepackage{amssymb}
\usepackage{fdsymbol}

\newcommand\blfootnote[1]{%
  \begingroup
  \renewcommand\thefootnote{}\footnote{#1}%
  \addtocounter{footnote}{-1}%
  \endgroup
}

\title{How to Train Your Fact Verifier: \\ Knowledge Transfer with Multimodal Open Models}

\author{
Jaeyoung Lee\textsuperscript{$\spadesuit$}\quad
Ximing Lu\textsuperscript{$\clubsuit\heartsuit$} \quad
Jack Hessel\textsuperscript{$\varheartsuit$} \quad
Faeze Brahman\textsuperscript{$\heartsuit$}  \\
\textbf{Youngjae Yu}\textsuperscript{$\blacklozenge$} \quad
\textbf{Yonatan Bisk}\textsuperscript{$\diamondsuit$} \quad
\textbf{Yejin Choi}\textsuperscript{$\clubsuit\heartsuit$} \quad
\textbf{Saadia Gabriel}\textsuperscript{$\dagger\ddagger$} \\[1ex]
\textsuperscript{$\spadesuit$}Seoul National University \quad
\textsuperscript{$\clubsuit$}University of Washington \quad
\textsuperscript{$\heartsuit$}Allen Institute for AI \quad \\
\textsuperscript{$\varheartsuit$}Samaya AI \quad
\textsuperscript{$\blacklozenge$}Yonsei University \quad
\textsuperscript{$\diamondsuit$}Carnegie Mellon University \\
\textsuperscript{$\dagger$}University of California, Los Angeles \quad
\textsuperscript{$\ddagger$}New York University
}

\begin{document}
\maketitle
\begin{abstract}
\blfootnote{Correspondence can be sent to: \texttt{jerry96@snu.ac.kr}, \texttt{yjy@yonsei.ac.kr} or \texttt{skgabrie@cs.ucla.edu}}
Given the growing influx of misinformation across news and social media, there is a critical need for systems that can provide effective real-time verification of news claims. Large language or multimodal model based verification has been proposed to scale up online policing mechanisms for mitigating spread of false and harmful content. While these can potentially reduce burden on human fact-checkers, such efforts may be hampered by foundation model training data becoming outdated. In this work, we test the limits of improving foundation model performance without continual updating through an initial study of knowledge transfer using either existing intra- and inter-domain benchmarks or explanations generated from large language models (LLMs). 

We evaluate on 12 public benchmarks for fact-checking and misinformation detection as well as two other tasks relevant to content moderation - toxicity and stance detection. Our results on two recent multi-modal fact-checking benchmarks, Mocheg and Fakeddit, indicate that knowledge transfer strategies can improve Fakeddit performance over the state-of-the-art by up to 1.7\% and Mocheg performance by up to 2.9\%. 
\end{abstract}

\section{Introduction}

\begin{figure}[t]
    \centering
    \includegraphics[width=0.5\textwidth, trim=4cm 0 0 0, clip]{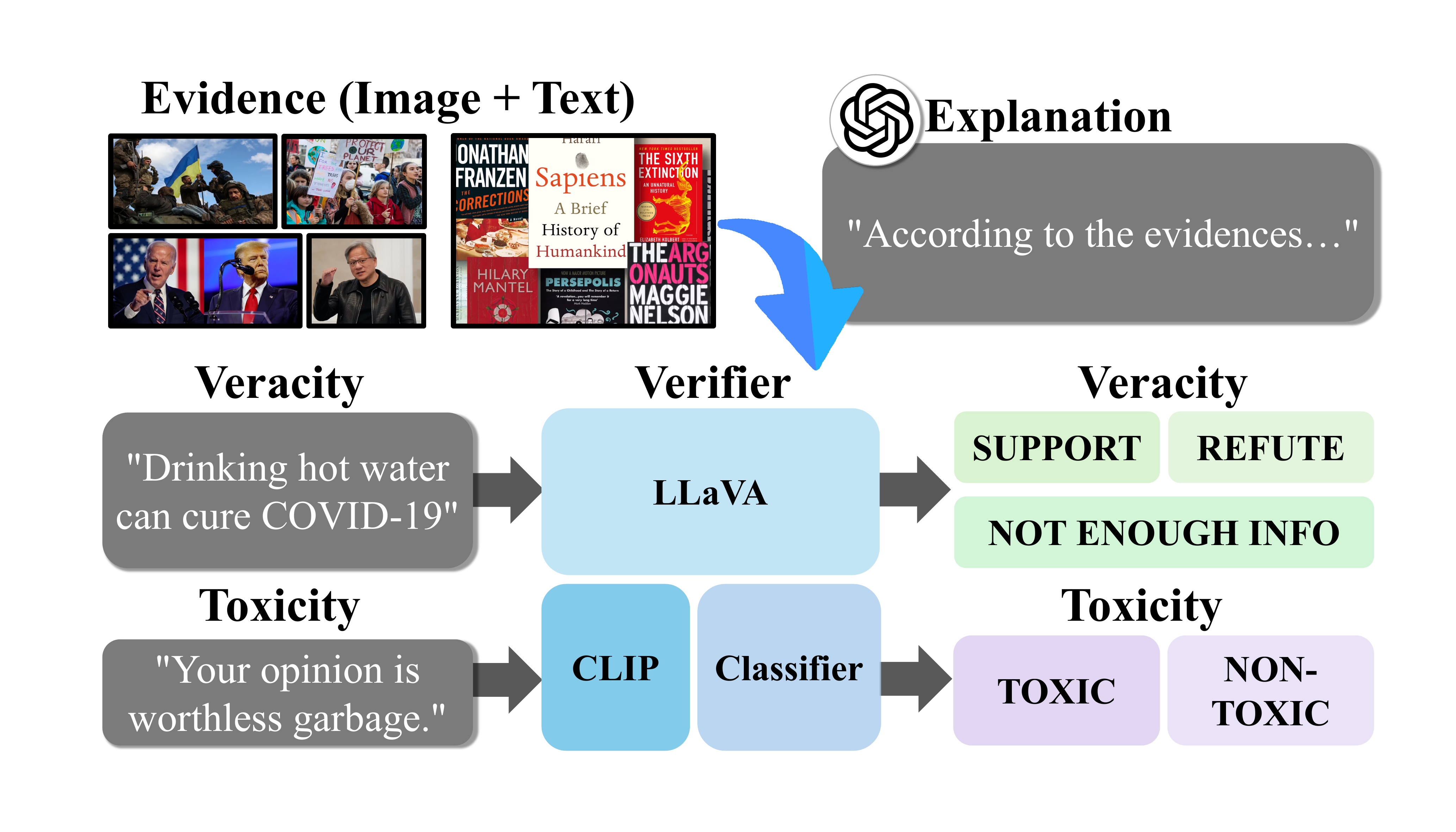}
    \caption{Visualization of our fact verification pipeline. Textual and visual evidence is embedded using a multimodal model, e.g. CLIP \cite{Radford2021LearningTV}, before passing through a classifier. Alternatively, a vision-language model such as LLaVA \cite{liu2024llavanext} can also be utilized. An external explanation generation model can be used to augment the input. Here we show a true claim from the Mocheg dataset. \cite{Yao2022EndtoEndMF}.} 
    \label{fig:umd}
\end{figure}

Top news stories rapidly go out-of-date and are replaced, e.g. during political election cycles. A recent study of Google Trends\footnote{https://www.newslifespan.com/} in 2018 found that popular news stories tend to stay relevant for a lifespan of only 7 days. The actual observed behavior of the general public seems incongruous with the current paradigm of automated fact-checking, which relies on static resources. Fact-checking organizations are increasingly turning to open pretrained language models like BERT \cite{devlin-etal-2019-bert} to scale up content moderation efforts \cite{wired-automated-factcheck,poynter-automated-factcheck}. These systems, trained on fixed knowledge bases, are not guaranteed to remain relevant as media narratives shift over time.

\begin{figure*}[t]
    \centering
    \includegraphics[width=\textwidth, trim=0 3.0cm 0 0, clip]{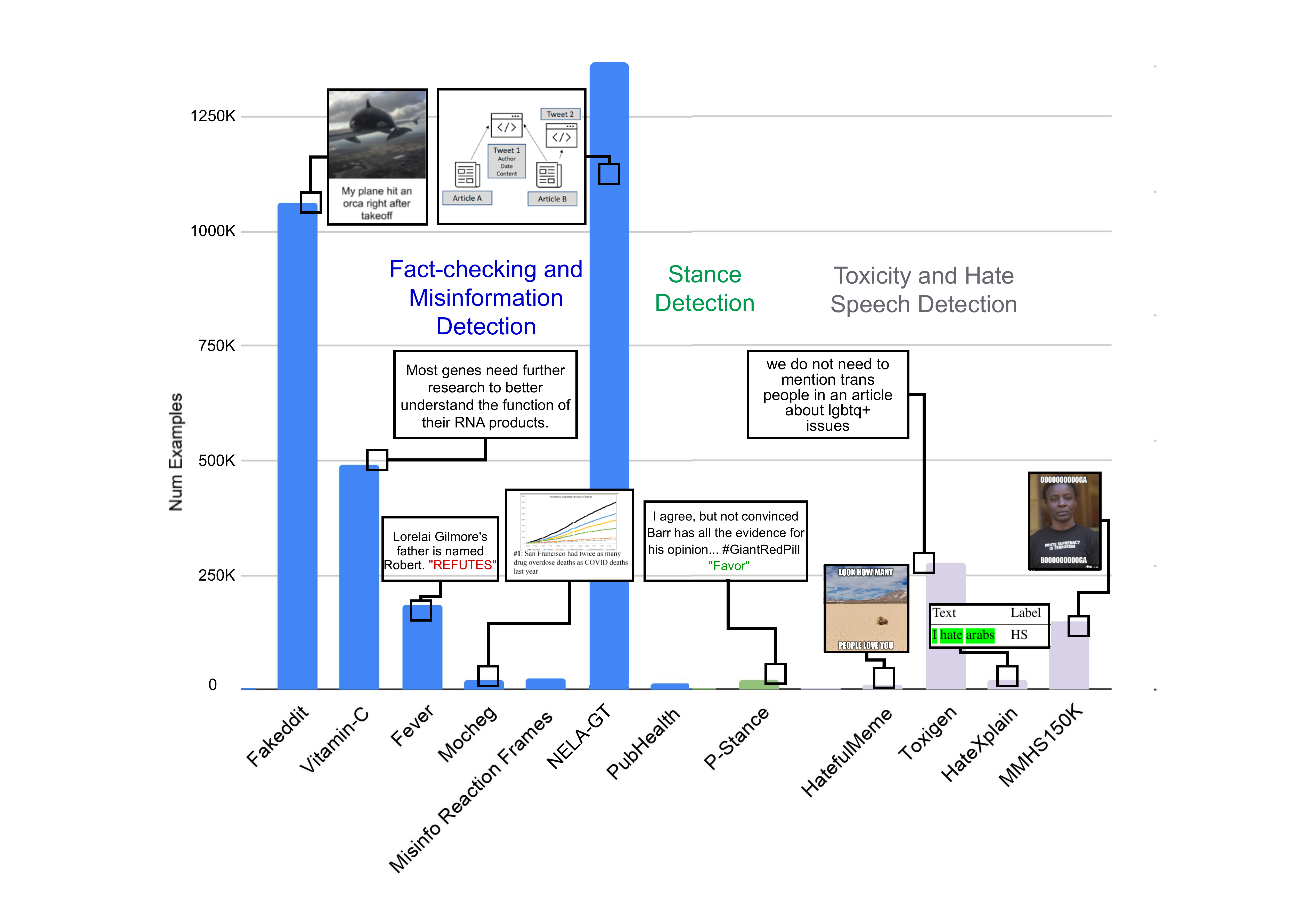}
    \caption{Finetuning and evaluation datasets considered in this work, with a breakdown by dataset scale and domain (misinformation detection, stance detection or toxicity detection). Fakeddit \cite{nakamura-etal-2020-fakeddit}, Mocheg \cite{Yao2022EndtoEndMF}, HatefulMemes \cite{Kiela2020TheHM} and MMHS150K \cite{Gomez2019ExploringHS} are all multi-modal datasets.}
    \label{fig:datasets}
\end{figure*}
%Also, while large language models (LLMs) have been conceptualized as knowledge bases \cite{petroni-etal-2019-language,West2021SymbolicKD,AlKhamissi2022ARO}, the veracity of information stored by neural language models may be too unreliable for fact-checking without filtering or interventions \cite{maynez-etal-2020-faithfulness,gabriel-etal-2021-go,West2021SymbolicKD}. This information is also brittle to an evolving world and may not generalize well to reasoning about current events \cite{10.1145/3442188.3445922,Hartvigsen2022AgingWG}.
 Motivated by previous approaches that have shown transformer-based models can effectively learn task-specific and linguistic reasoning skills from unified pretraining, e.g. for QA \cite{khashabi-etal-2020-unifiedqa} and small-scale text-only misinformation detection \cite{Lee2021OnUM}, we conduct a large-scale study of transfer learning across diverse text-only and multimodal datasets to boost reasoning capabilities of misinformation detection systems. We consider \textit{intra-domain transfer} from 6 misinformation detection and fact-checking datasets, as well as \textit{inter-domain transfer},  where a fact verification model is jointly trained across other content moderation tasks (e.g. hate speech detection) to overcome brittleness to biases of fact-checking datasets. We also consider the impact of parametric knowledge transfer from larger-scale closed models like GPT-3.5-turbo \cite{GPT35} and GPT-4 \cite{Achiam2023GPT4TR} through generated explanations for a specific veracity label. We also address how transfer learning impacts a model generalization issue known as \textit{covariate shift} \cite{SHIMODAIRA2000227}, when the test data distribution diverges from the training distribution despite consistency in labeling procedures. A very tangible example of this is when evaluated news topics change entirely from training data due to a major unforeseen event like the
 Covid-19 pandemic.
 
Given a claim like \textit{``The FBI warned that smart TVs can `spy' on their owners,''} Figure \ref{fig:umd} shows how our full verification pipeline can be used to predict the reliability of a social media claim. The system consists of (1) a unified fact verification model $M_{FC}$ that predicts the veracity or potential harm (e.g. toxicity) of multi-modal inputs, and (2) an explanation model $M_{EG}$ prompted to generate explanations of claim veracity. Following from theories of human cognition and language interpretation, which relies not only on continuous learning of facts but on commonsense world understanding \cite[e.g,][]{Newell1973HumanPS,Fillmore1976FRAMESA}, as well as recent work showing the effectiveness of LLM explanations for countering misinformation \cite{Hsu2023IsET, Chen2023CombatingMI, Wan2024DELLGR, Gabriel2024Generative}, we seek to supervise verifier training with examples of correct and noisy fact verification reasoning using $M_{EG}$. 

\paragraph{Conclusions \& Resources.} We evaluate our verification pipeline on 12 existing text-only and multi-modal fact verification, hate speech and stance detection benchmarks. Our results confirm that fact verification models trained on common benchmarks are extremely brittle to distribution shift and indicate data diversity is an important factor in high-performing fact verifiers over scale alone.  Our best intra-domain mixture improves Fakeddit results by 1.7\% F1, leading to a performance of 93.42 F1. We also find that knowledge distillation through GPT-4o and GPT-3.5-turbo explanations can boost Mocheg performance over the state-of-the-art by 2.9\%. Beyond fact-checking, our study has implications for other important content moderation domains like hate speech detection, where we find knowledge transfer from fact-checking data can boost performance by 13.65\%. We will release our code, pretrained models and evaluation data to encourage further research on robust fact verification.

\section{Motivation \& Related Work}

\paragraph{Automated Fact-Checking}

Given the rapid proliferation of misinformation on social media, there is a critical need for automatic tools that can assist human fact-checkers \cite{Nakov2021AutomatedFF}. Much of the earlier work in this area relied on linguistic cues or social media network features \cite[e.g.,][]{Wang2017LiarLP,rashkin-etal-2017-truth,PrezRosas2017AutomaticDO,Yang2019UnsupervisedFN}. Later work on detection of mis- and disinformation has considered transformer-based approaches, notably \cite{zellers2019neuralfakenews}. Most similar to our work is \cite{Lee2021OnUM}, which also does unified misinformation detection on a much smaller scale by training a model on various unimodal misinformation detection corpora. Seperately, a body of prior work also explores fact-checking explanations \cite[e.g.,][]{atanasova-etal-2020-generating-fact}, and prior to us, \citet{angeli-manning-2014-naturalli} draws a connection between claim verification and the natural logical inference underlying commonsense acquisition. 

\paragraph{Robustness in Fact-Checking}
Prior work has highlighted challenges to automated fact-checking like insufficient evidence \cite{atanasova-etal-2022-fact} or spurious correlations used for evidence retrieval \cite{asai-etal-2022-evidentiality}. We address these knowledge gaps by exploring the use of machine-generated explanations to aid in fact verification. This is similar to the motivation behind the VITAMIN-C dataset \cite{schuster-etal-2021-get}, however VITAMIN-C only considers Wikipedia revisions and is not multi-modal. Most recently, \cite{Caramancion2023NewsVS,Cao2023AreLL,guan2023language} have delved into the limitations of LLMs for fact verification, noting that while the ability of LLMs to verify inputs surpasses their ability to generate factual content, they are still far less reliable than human fact-checkers. 

\section{Transfer Learning Strategies}

In this section, we first describe the basic experimental setup for multimodal fact-checking. We then describe transfer learning methodologies based on whether knowledge is being distilled from more diverse sources within the same domain, transfer of knowledge across domains (e.g. misinformation detection and toxicity detection), or from parametric knowledge through LLM-generated explanations. 

%examine the interplay of dataset dynamics (e.g. label consistency, transferability of generalizable reasoning abilities), in order to determine the optimal data mixture for unified pretraining of our verification model.  

\subsection{Base Verification Architecture}

For our base verification model $M_{FC}$ we use the vision-and-language classification model introduced by \cite{Yao2022EndtoEndMF}. This model jointly encodes a textual claim, text evidence and image evidence using CLIP-base \cite{Radford2021LearningTV}. This representation is then used for the intermediate task of predicting the stance of the evidence towards the claim. The output stance representations are aggregated to predict the claim veracity using a final linear classification layer. We modify the base architecture with several larger embedding models: CLIP-large, CLIP-large-336 and 7B LLaVA-NeXT \cite{liu2024llavanext} which we will refer to as LLaVA.

\subsection{Dataset Mixtures}

Figure \ref{fig:datasets} shows the 12 datasets considered in this work, along with their 3 respective domains (fact checking / misinformation detection, toxicity / hate speech detection and stance detection).\footnote{For Mocheg, we use the original training/val/test splits described in \url{https://arxiv.org/abs/2205.12487v1}.} \\ For intra-domain analysis, misinformation and fact-checking datasets are normalized to have a shared label space (\textit{supported}, \textit{refuted}, \textit{nei}). All results are reported using gold image and text evidence. 

%delineate task-specific data by appending ``[task]" to the claim text. 

\begin{figure*}
\centering
\begin{subfigure}{.5\textwidth}
  \centering
  \includegraphics[width=1\linewidth]{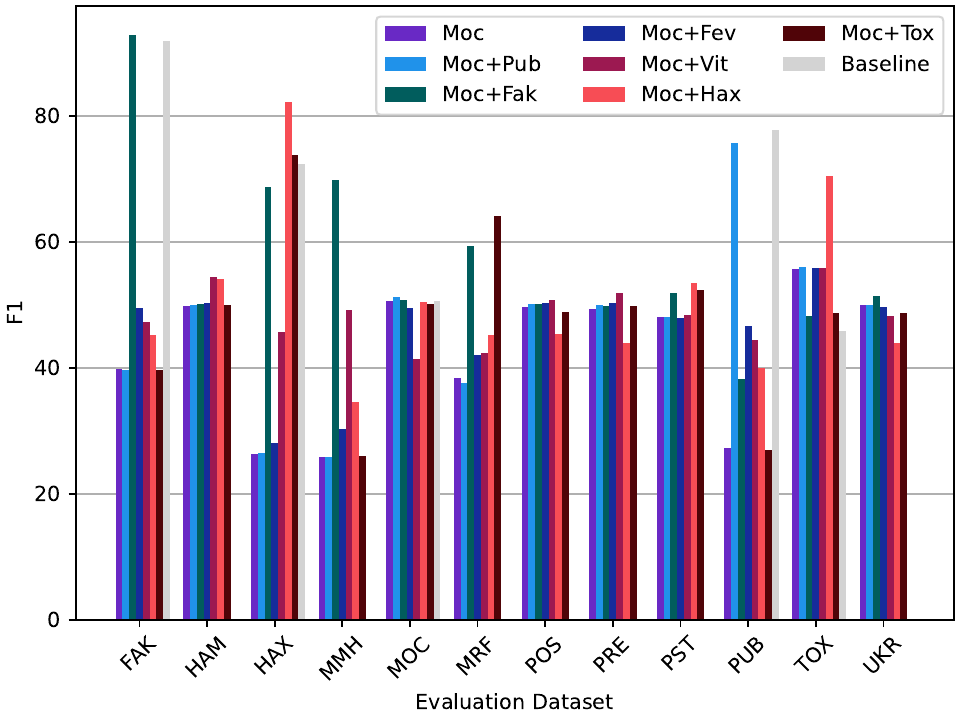}
  \caption{2-dataset mixtures, Mocheg. }
  \label{fig:2_mocheg}
\end{subfigure}%
\begin{subfigure}{.5\textwidth}
  \centering
  \includegraphics[width=1\linewidth]{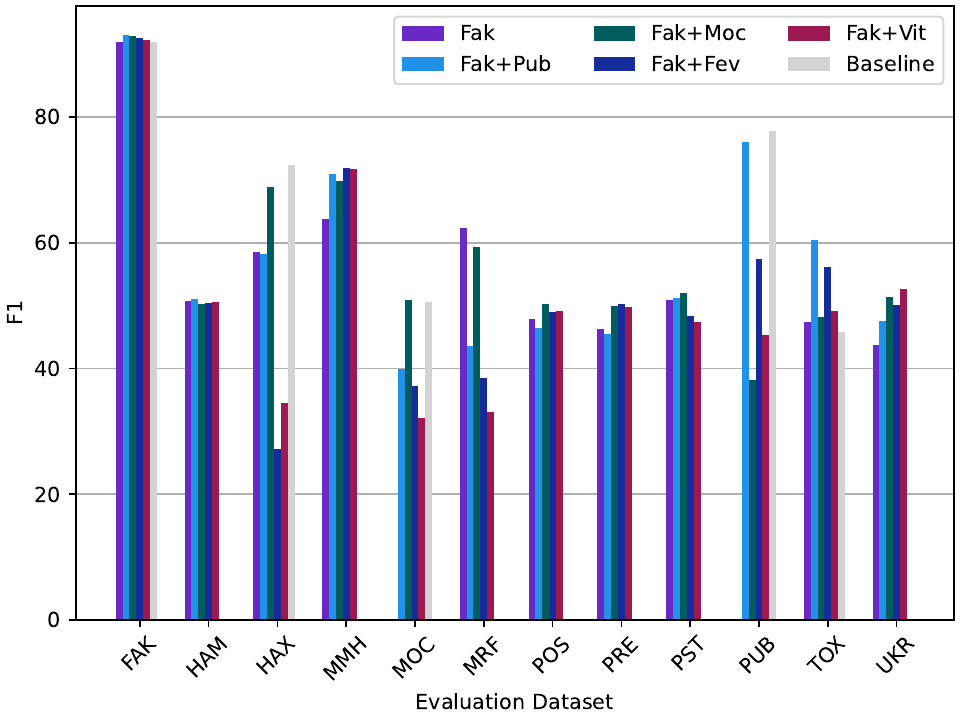}
  \caption{2-dataset mixtures, Fakeddit.}
  \label{fig:2_fakeddit}
\end{subfigure}
\caption{Transfer learning results for both intra- and inter-domain 2-dataset mixtures used to train our largest multimodal model (CLIP-large-336). Due to the large size and computational demands of Fakeddit, we only evaluate intra-domain mixtures. Baseline results are shown for Fakeddit, Hateful Memes, Mocheg, PubHealth and Toxigen eval sets in light gray.}
\label{fig:results_2}
\end{figure*}

\subsection{LLM Explanations}
\label{sec:explain}

For the explanation generation model $M_{EG}$, we use GPT-3.5-turbo or GPT-4o \cite{Achiam2023GPT4TR} with text-only inputs. We instruct the model to provide an explanation for why a claim $x$ has a specific label $y$ using the following system and user prompts:

\begin{quote}
    \textbf{System Prompt}: 
    
\textit{You are an AI assistant skilled in fact-checking. Your role is to generate justifications for relationships between claims and evidence. Analyze the information provided and explain why the evidence supports or refutes the claim based on the labeled relationship.}
\end{quote}

\begin{quote} 
    \textbf{User Prompt}: 

\textit{Here is the information:}

\textit{Claim: {\{claim\}}}

\textit{Evidence: {\{evidence\}}}

\textit{Relationship: {\{label\}}}
\newpage

\# \textit{Task}

\textit{Please generate a explanation that justifies the specified relationship between the claim and the evidence}

\# \textit{Requirements} 

- \textit{You should provide explanation without expressing the relationship explicitly.}

- \textit{You should be concise and clear.}

- \textit{The answer should be less than 100 words.}
\end{quote}

%consider two variations: an explanation type we call ``\textit{Refuted}" in which we always assume the claim is false, and an explanation ``\textit{Label}" in which we may infer any given label type. For ``\textit{Refuted}", we use 3-shot demonstration prompting with ruling statements from the Mocheg dataset as a template for an explanation.  Explanations are generated using GPT-3.5-Turbo. In the prompt instruction, we can either explicitly constrain the model to only use commonsense knowledge over retrieved evidence or use both types of knowledge. An example prompt for the ``\textit{Label}" explanation type is shown below:\footnote{The same knowledge type instruction is used for ``\textit{Refuted}", but here we specify the claim must be false.}
%\begin{quote}
%    \textit{In 100 words or less, argue step-by-step for why a claim may be [label] using commonsense knowledge only and not retrieved factual knowledge...}
%\end{quote}

On the Mocheg dataset, GPT-3.5-turbo achieves 45.17 F1 and GPT-4o achieves 65.85 F1. Despite the strong performance of GPT-4o, we include the older GPT-3.5-turbo model in experimentation given that GPT-4o's predictive abilities may be partly explained by dataset leakage of Mocheg in GPT-4's training set.

%three variations: one with full access to parametric knowledge in which we prompt GPT-3.5-Turbo with an explanation instruction and claim, one in which we control access to recent data by using a distilled pretrained model based on GPT-2 XL, and the same distilled GPT-2 XL model without any pretraining.

\section{Experimental Setup}

\subsection{Model Training}

\begin{figure*}[t]
    \centering
    \includegraphics[width=.9\textwidth]{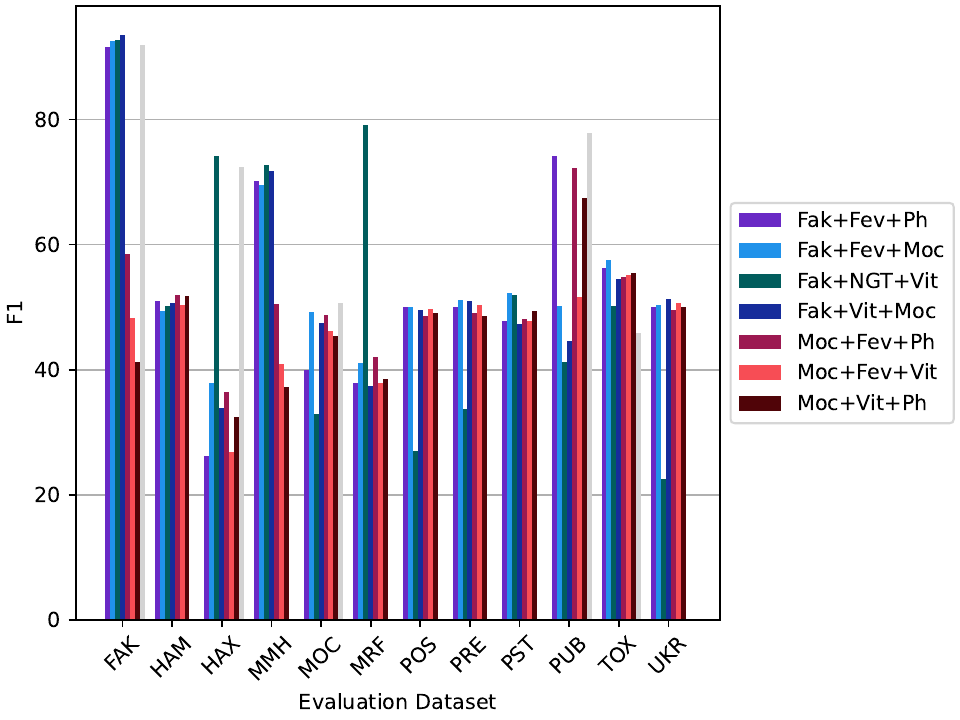}
    \caption{Transfer learning results for intra-domain 3-dataset mixtures used to train our largest multimodal CLIP model (CLIP-large-336).}
    \label{fig:results_3}
\end{figure*}

We used publicly open models from Huggingface. Specifically, (1) openai/clip-vit-base-patch32 , (2) openai/clip-vit-large-patch14, (3) openai/clip-vit-large-patch14-336, and (4) llava-hf/llava-v1.6-mistral-7b-hf were used.

\paragraph{CLIP-based Verifiers.} All the CLIP-based verifiers were trained with 2048 batch size using grad-accumulation with minibatch size of 256. The models were trained with 50 epochs without early stopping. Adam optimizer was used. The learning rate was 1e-3. Models were trained on RTX3090, RTX4090, RTX8000, A6000, A100, and H100 GPUs.

\paragraph{LLaVA-based Verifier.} LLaVA models were trained using LoRA(rank=64, lora alpha=16, dropout=0.05), targeting all linear layers. For precision, bfloat16 was adopted. The max sequence length was 2048. The batch size was 32, using grad accumulation with the minibatch size of 1. AdamW was used for the optimizer, with the learning rate of 2e-5. For the scheduler, cosine-annealing learning rate scheduler was used. Models were trained on A100 and H100.

\paragraph{Other Details} All datasets are primarily in English and publicly available for research purposes. For all results, we report a single run. We use most recent version of NLTK and SciPy packages for pre-processing and analysis.

\section{Evaluation Details} 

We evaluate claim verification, toxicity detection and stance detection performance using F1. For three-label dataset mixtures evaluated on two-label datasets, we map labels 0 and 1 between the datasets (e.g. supports $\rightarrow$ benign, and refutes $\rightarrow$ toxic). If the predicted label is 2, we determine $\hat{y}$ for the two-label dataset from the model output probability distribution by finding the most probable label from {0,1}. We consider 12 evaluation sets: Fakeddit (fak), HatefulMemes (ham), HateXplain (hax), MMHS150K (mmh), Mocheg (moc), Misinfo Reaction Frames (mrf), 3 temporal Nela-GT subsets (pre, pos, ukr) described in the next section, P-Stance (pst), PubHealth (ph), and Toxigen (tox). 

\subsection{Temporal Shift Evaluation}

To test the robustness of misinformation detection systems to temporal shift, we consider three different filters on the NELA-GT 2020-2022 dataset \cite{gruppi2021nelagt2020,gruppi2023nelagt2022} which can be used to partition eval examples as in-distribution or out-of-distribution depending on the training cut-off of the underlying model. We relate these partitions to notable recent historical events, with \textit{pre} representing news before the first vaccine release in December 2020, \textit{pos} representing news from 2021 after the vaccine release, and \textit{ukr} representing recent news from 2022 relating to the Ukraine-Russia war. Each evaluation set contains 1000 randomly sampled claims from the relevant subsets of NELA-GT. 

\section{Learning from Dataset Mixtures}

In Figures \ref{fig:results_2} and \ref{fig:results_3},\footnote{We omit results from Fakeddit on Mocheg and PubHealth in Figure \ref{fig:2_fakeddit} since Fakeddit lacks nei labels.} we show that the transfer learning mixture substantially impacts the in-distribution and out-of-distribution performance of CLIP-large-336 trained on fact-checking and toxicity detection benchmarks. While some mixtures (e.g. \textit{Mocheg + Fakeddit + PubHealth}) actually decrease fact-checking benchmark performance of multimodal models, others (e.g. \textit{Mocheg + PubHealth} for Mocheg, \textit{Mocheg + Vitamin-C  + Fakeddit} for Fakeddit) improve by up to 1.56 F1 over single dataset baseline results. We also find that fact-checking mixtures can lead to strong performance at hate speech detection (up to 72.75 F1 for MMHS and 82.22 F1 for HateXplain). In the next few sections, we discuss in-depth how model performance and generalization capabilities are affected by our dataset mixtures.  

\subsection{Learning within domain}

Shown in Figures \ref{fig:results_2} and  \ref{fig:results_3}, fact-checking models are particularly brittle to domain shift. Dataset mixtures get close to random performance on temporal evaluation sets \textit{pre}, \textit{pos} and \textit{ukr}. 

\paragraph{Improvements from PubHealth Mixtures.} For PubHealth, no mixtures improve over the single dataset baseline (77.81 F1), though including PubHealth in 2-dataset mixtures improves performance for other fact-checking/misinformation evaluation sets: Fakeddit (+1.09 F1), Mocheg (+0.61 F1). Figure \ref{fig:2_mocheg} shows that Fakeddit + Mocheg consistently outperforms or is comparable to other out-of-distribution mixtures. 

\paragraph{Data Diversity Over Scale.} Notably, from Figure \ref{fig:2_fakeddit} Fakeddit does not generalize well on its own despite being significantly larger than other training datasets (1,063,106 samples), indicating there is still benefit from data diversity introduced by mixtures over a single large-scale dataset. 

\paragraph{In-distribution Performance.} For fact-checking, unsurprisingly the strongest performance on a given eval set comes from mixtures including the associated training set. Interestingly, this is not always true for toxicity/hate speech benchmarks. We discuss this in the next section, and full results can be found in the Appendix. 

\begin{table*}[t]
\parbox{.45\linewidth}{
\centering

  \begin{tabular}{l|c|c|c|c}
  \specialrule{1.5pt}{-1.5pt}{0pt}
  & \multicolumn{2}{c}{\textbf{CLIP-base}} & \multicolumn{2}{c}{\textbf{CLIP-large}} \\ 
      Scenario & $\Delta F1_{4o}$ & $\Delta F1_{3.5}$ & $\Delta F1_{4o}$ & $\Delta F1_{3.5}$ \\ \toprule     
      Random  & + 1.39 & + 0.29 & + 0.45 & + 0.17 \\      
      Opposite &  + 0.16 & + 4.38 & + 1.48 & + 1.03 \\ 
      Always Supports &  + 2.79 & - 1.02 & + 1.60 &  - 0.69\\
      Always Refutes  &  + 1.39 & + 0.21 & - 0.90 & + 0.04\\     
      Always NEI &  + 2.87  & - 0.20 & - 0.86 &  + 1.72\\  
      All & + 1.39 & - 0.78 & - 0.69  & - 0.86\\
      Guided & + 1.84 & + 1.35 & + 1.15 & + 0.41 \\
      Oracle & + 10.03 & + 11.51 & + 10.89 & + 12.49 \\ \specialrule{1.5pt}{-1.5pt}{0pt}  
 \end{tabular}
 }\hfill 
 \parbox{.45\linewidth}{
 \centering
   \begin{tabular}{l|c|c}
   \specialrule{1.5pt}{-1.5pt}{0pt}
   & \multicolumn{2}{c}{\textbf{LLaVA}} \\   
       Scenario & $\Delta F1_{4o}$ & $\Delta F1_{3.5}$  \\ \toprule
       Random & - 1.35 & - 1.03 \\
       Guided & + 1.84 & - 0.70  \\
       Oracle &  + 30.38 & + 23.67\\ \specialrule{1.5pt}{-1.5pt}{0pt}
    \end{tabular}
 }
  \caption{Mocheg test F1 after augmenting the training set with LLM explanations. $\Delta F1_{4o}$ denotes the difference between F1 results from the GPT-4o augmented model and the baseline F1, while $\Delta F1_{3.5}$ denotes the difference from the GPT-3.5-turbo augmented model and the baseline model. Here we can consider Oracle as an upperbound on performance. The left table shows CLIP results and the right table shows LLaVA results.}
  \label{tab:results_explain}
\end{table*}

\subsection{Learning across domains}

From Figure \ref{fig:results_2}, we do not find that inter-domain knowledge transfer from toxicity/hate speech detection aids in fact-checking. This indicates that it may be most critical for future work to focus on development of diverse, high-quality fact-checking datasets rather than cross-task learning. However, inter-domain transfer from fact-checking datasets does improve toxicity/hate speech detection. For example, the Mocheg + HateXplain mixture improves HateXplain performance by 13.65\% and also performs better than HateXplain alone (by 9.97\%) on Toxigen, another hate speech detection dataset. 

%In general, transfer learning proved ineffective across fact-checking and misinformation datasets with the exception of PubHealth (+1.34 F1).  

\subsection{Discussion}

For both intra- and inter-domain mixtures, discrepancies in the structure and label space of the datasets may harm generalization capabilities. For example, we observe that adding evidence-less datasets to mixtures including fact-checking datasets with \textit{nei} labels leads to over-prediction of \textit{nei}. A potential solution is dataset normalization, but this requires evidence retrieval. Our next section is motivated by this issue, where we explore effects of knowledge transfer from GPT-4o generated veracity explanations on Mocheg performance. This could be used to close the gap between evidence-augmented and evidence-less datasets. 

\section{Learning from Explanations}

In the next section, we explore the use of GPT-4o and GPT-3.5-turbo explanations as silver evidence. 

\subsection{Explanation Scenarios}

We consider 8 different scenarios which assume different levels of access to gold labels, and compare against baseline results without explanations. We evaluate with the following scenarios that assume knowledge of the gold label:
\begin{itemize}
    \item \textbf{Oracle}: We use the gold claim labels for generating explanations. This acts as an upper-bound on explanation performance.  
    \item \textbf{Opposite}: We generate explanations arguing for the opposite label to the gold label if the gold label is true/false. This tests the model's ability to learn from contradictory information. 
\end{itemize}

The next 6 scenarios are more realistic, and do not assume knowledge of the gold claim labels: 

\begin{itemize}
    \item \textbf{Random}: We generate explanations using randomly selected labels for each claim. 
    \item \textbf{All}: We use explanations for all three possible claim labels. 
    \item \textbf{Always Supports}: We always use \textit{supports} as the label to generate explanations.  
    \item \textbf{Always Refutes}: We always use \textit{refutes} as the label to generate explanations.  
    \item \textbf{Always NEI}: We always use \textit{nei} as the label to generate explanations.  
    \item \textbf{Guided}: We use the explanation generation model's zero-shot predicted claim labels to generate explanations. 
\end{itemize}

\subsection{Verification Results}

Table \ref{tab:results_explain} shows effect on baseline Mocheg results (49.18 F1 for CLIP-base, 50.49 F1 for CLIP-large and 63.23 F1 for LLaVA) from training models on the Mocheg dataset and GPT4-o generated explanations. The smaller CLIP models always benefit from guided explanations with improvements of 0.41-1.84 F1, even though the accuracy of GPT-3.5-turbo is lower than Mocheg finetuned CLIP-large. We note that explanations generated using random labels also improve results slightly. Results for explanations generated using a single label are mixed, and will potentially be influenced by label distributions. Using explanations generated for every label was generally not beneficial, possibly because of information being ignored when considering the longer context.

Following our findings with CLIP, we see if these performance gains hold for LLaVA, a recent state-of-the-art vision-and-language model. While LLaVA results do not improve using GPT-3.5-turbo explanations, we do find that GPT-4o explanation augmented models consistently outperform the LLaVA baseline, leading to state-of-the-art open model results (65.07 F1) on Mocheg.\footnote{To the best of our knowledge.} 

\subsection{Quality of Explanations}

\paragraph{Evaluation Setup.} We conduct a human evaluation on the Amazon Mechanical Turk crowdsourcing platform \footnote{\url{https://www.mturk.com/}} to assess how multimodal models may learn from GPT-4o and GPT-3.5-turbo generated explanations. We randomly sample 36 claims from the Mocheg eval set and consider pairs of GPT-4o and GPT-3.5-turbo explanations arguing for the same random label (\textit{supports}, \textit{refutes} or \textit{nei}) for each claim. We then ask participants the following 6 questions to measure the reasoning capabilities of the two explanation generation models: 

\begin{itemize}[noitemsep,leftmargin=15pt]
    \item \textbf{Q1}: Does the AI explanation use inaccurate reasoning to prove the claim is true or false? (yes/no)
    \item \textbf{Q2}: Does the AI explanation use commonsense reasoning to prove the claim is true or false? (yes/no) 
    \item \textbf{Q3}: Does the AI explanation use knowledge of specific events to prove the claim is true or false? (yes/no)
    \item \textbf{Q4}: Does the AI explanation use domain knowledge (e.g. scientific or legal knowledge) to prove the claim is true or false? (yes/no)
    \item \textbf{Q5}: Can you predict the claim label from the explanation? (the label is true, the label is false, the label is unprovable, no)
    \item \textbf{Q6}: What is the overall quality of the explanation ? (1-5 scale)
\end{itemize}

We explicitly instruct the participants that they can make use of external sources (e.g. Google Search) when verifying explanation quality. Each explanation is assessed by 5 participants, who have prior experience working on misinformation detection tasks and have passed basic attention checks. We filter annotators who contribute to low inter-annotator agreement based on Fleiss' $\kappa$ for judgments of explanation quality. We observe fair agreement of Fleiss' $\kappa = 0.276$ for GPT-4 explanation quality. For each claim and evaluation question except predictability, we take a majority vote over participant responses to determine the answer. For predictability, we check if at least 1 participant could identify the correct explanation label. 
\begin{figure}
    \centering
    \includegraphics[width=1\linewidth]{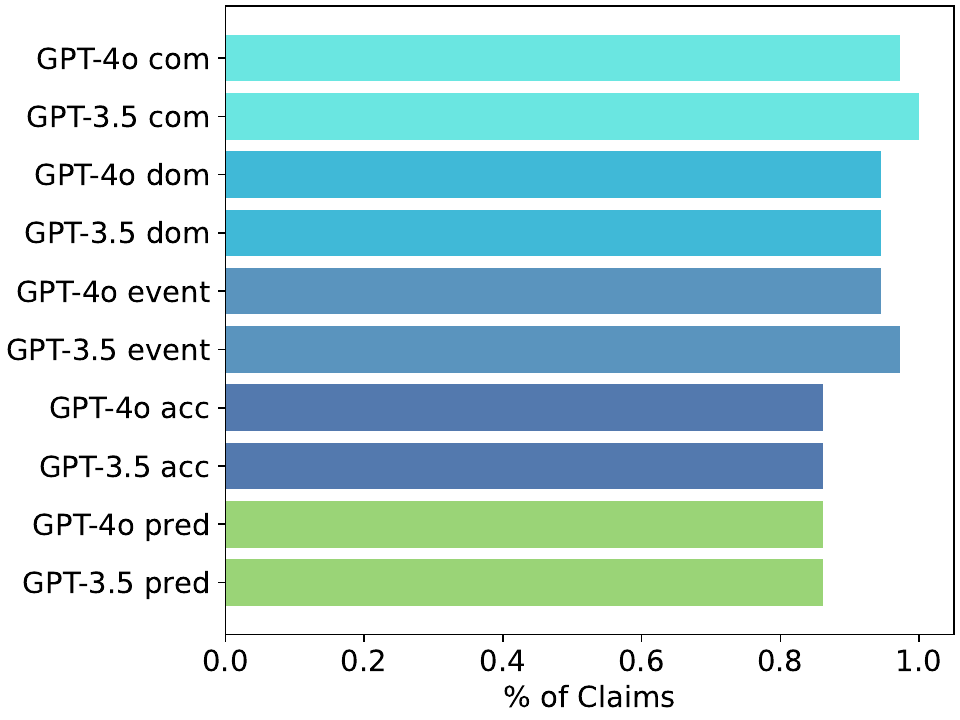}
    \caption{Human evaluation questions 1-5 for explanation quality. We measure use of commonsense reasoning (com), use of domain knowledge (dom), use of event knowledge (event), accuracy of reasoning (acc) and label predictability (pred).}
    \label{fig:ex_human_eval}
\end{figure}
\paragraph{Results.} As shown by Figure \ref{fig:ex_human_eval}, explanations use a combination of commonsense reasoning and domain-specific knowledge, as well as evidence retrieval from known events. Explanations were generally found to be accurate (86.11\% of explanations for both models). Participants struggled to predict the label used to generate a given explanation, with the majority of participants able to predict the correct label in less than half of cases. When we consider cases where at least one participant could identify the correct explanation label, we find that GPT-4o explanation labels and GPT-3.5-turbo explanations have equal predictability. Quality is comparable between the models (4.00 for GPT-4o vs. 4.04 for GPT-3.5-turbo). 

%During training, we have access to gold labels in order to construct an argument for or against a particular claim. However at inference, we must assume a particular label for the explanation. We test a scenario in which the explanation always argues against the claim and find this leads to slight improvement (+1.14 F1).  

\section{Conclusion}

In conclusion, we test three different transfer learning strategies to measure impact on multimodal fact-checking performance and improve results on two widely used benchmarks. We also discuss future steps for robust fact verification, such as using closed LLMs to expand reasoning capabilities of open foundation models. We show that explanations generated from powerful LLMs like GPT-4o and GPT-3.5-turbo can boost performance of smaller models. 

\section{Ethics Statement \& Limitations}

Given our findings, we urge caution in selection of fact verification models. Even more so than other content moderation domains like hate speech detection, our study suggests that strong in-distribution performance of fact verifiers is not indicative of strong general performance. Users and researchers should bear this in mind when deploying out-of-the-box fact verifiers on unseen data. 

While we provide a preliminary study on knowledge transfer from dataset mixtures and LLM-generated explanations, future work may expand upon this by considering an even broader range of datasets and tasks. Another future direction may be use of visual information in explanation generation. 

\section{Acknowledgements}

We thank colleagues at UW, AI2 and NYU for thought-provoking discussions that contributed to this work, specifically Kate Starbird, Chandra Bhagavatula and He He. We also thank OpenAI for providing credits to access models.

\bibliography{anthology,custom}

\begin{thebibliography}{36}
\expandafter\ifx\csname natexlab\endcsname\relax\def\natexlab#1{#1}\fi

\bibitem[{Abels(2022)}]{poynter-automated-factcheck}
Grace Abels. 2022.
\newblock \href
  {https://www.poynter.org/fact-checking/2022/how-will-automated-fact-checking-work/}
  {What is the future of automated fact-checking? fact-checkers discuss.}
\newblock \emph{Poynter}.

\bibitem[{Angeli and Manning(2014)}]{angeli-manning-2014-naturalli}
Gabor Angeli and Christopher~D. Manning. 2014.
\newblock \href {https://doi.org/10.3115/v1/D14-1059} {{N}atural{LI}: Natural
  logic inference for common sense reasoning}.
\newblock In \emph{Proceedings of the 2014 Conference on Empirical Methods in
  Natural Language Processing ({EMNLP})}, pages 534--545, Doha, Qatar.
  Association for Computational Linguistics.

\bibitem[{Asai et~al.(2022)Asai, Gardner, and
  Hajishirzi}]{asai-etal-2022-evidentiality}
Akari Asai, Matt Gardner, and Hannaneh Hajishirzi. 2022.
\newblock \href {https://doi.org/10.18653/v1/2022.naacl-main.162}
  {Evidentiality-guided generation for knowledge-intensive {NLP} tasks}.
\newblock In \emph{Proceedings of the 2022 Conference of the North American
  Chapter of the Association for Computational Linguistics: Human Language
  Technologies}, pages 2226--2243, Seattle, United States. Association for
  Computational Linguistics.

\bibitem[{Atanasova et~al.(2020)Atanasova, Simonsen, Lioma, and
  Augenstein}]{atanasova-etal-2020-generating-fact}
Pepa Atanasova, Jakob~Grue Simonsen, Christina Lioma, and Isabelle Augenstein.
  2020.
\newblock \href {https://doi.org/10.18653/v1/2020.acl-main.656} {Generating
  fact checking explanations}.
\newblock In \emph{Proceedings of the 58th Annual Meeting of the Association
  for Computational Linguistics}, pages 7352--7364, Online. Association for
  Computational Linguistics.

\bibitem[{Atanasova et~al.(2022)Atanasova, Simonsen, Lioma, and
  Augenstein}]{atanasova-etal-2022-fact}
Pepa Atanasova, Jakob~Grue Simonsen, Christina Lioma, and Isabelle Augenstein.
  2022.
\newblock \href {https://doi.org/10.1162/tacl_a_00486} {Fact checking with
  insufficient evidence}.
\newblock \emph{Transactions of the Association for Computational Linguistics},
  10:746--763.

\bibitem[{Cao et~al.(2023)Cao, Wei, Chen, Zhou, and Hu}]{Cao2023AreLL}
Han Cao, Lingwei Wei, Mengyang Chen, Wei Zhou, and Song Hu. 2023.
\newblock \href {https://api.semanticscholar.org/CorpusID:265499074} {Are large
  language models good fact checkers: A preliminary study}.
\newblock \emph{ArXiv}, abs/2311.17355.

\bibitem[{Caramancion(2023)}]{Caramancion2023NewsVS}
Kevin~Matthe Caramancion. 2023.
\newblock \href {https://api.semanticscholar.org/CorpusID:259308983} {News
  verifiers showdown: A comparative performance evaluation of chatgpt 3.5,
  chatgpt 4.0, bing ai, and bard in news fact-checking}.
\newblock \emph{ArXiv}, abs/2306.17176.

\bibitem[{Chen and Shu(2023)}]{Chen2023CombatingMI}
Canyu Chen and Kai Shu. 2023.
\newblock \href {https://api.semanticscholar.org/CorpusID:265128809} {Combating
  misinformation in the age of llms: Opportunities and challenges}.
\newblock \emph{ArXiv}, abs/2311.05656.

\bibitem[{Devlin et~al.(2019)Devlin, Chang, Lee, and
  Toutanova}]{devlin-etal-2019-bert}
Jacob Devlin, Ming-Wei Chang, Kenton Lee, and Kristina Toutanova. 2019.
\newblock \href {https://doi.org/10.18653/v1/N19-1423} {{BERT}: Pre-training of
  deep bidirectional transformers for language understanding}.
\newblock In \emph{Proceedings of the 2019 Conference of the North {A}merican
  Chapter of the Association for Computational Linguistics: Human Language
  Technologies, Volume 1 (Long and Short Papers)}, pages 4171--4186,
  Minneapolis, Minnesota. Association for Computational Linguistics.

\bibitem[{Fillmore(1976)}]{Fillmore1976FRAMESA}
Charles~J. Fillmore. 1976.
\newblock Frame semantics and the nature of language *.
\newblock \emph{Annals of the New York Academy of Sciences}, 280.

\bibitem[{Gabriel et~al.(2024)Gabriel, Lyu, Siderius, Ghassemi, Andreas, and
  Ozdaglar}]{Gabriel2024Generative}
Saadia Gabriel, Liang Lyu, James Siderius, Marzyeh Ghassemi, Jacob Andreas, and
  Asu Ozdaglar. 2024.
\newblock Generative {AI} in the {Era} of '{Alternative} {Facts}'.
\newblock \emph{An MIT Exploration of Generative AI}.
\newblock Https://mit-genai.pubpub.org/pub/cnks7gwl.

\bibitem[{Gomez et~al.(2019)Gomez, Gibert, G{\'o}mez, and
  Karatzas}]{Gomez2019ExploringHS}
Raul Gomez, Jaume Gibert, Llu{\'i}s G{\'o}mez, and Dimosthenis Karatzas. 2019.
\newblock Exploring hate speech detection in multimodal publications.
\newblock \emph{2020 IEEE Winter Conference on Applications of Computer Vision
  (WACV)}, pages 1459--1467.

\bibitem[{Gruppi et~al.(2021)Gruppi, Horne, and Adalı}]{gruppi2021nelagt2020}
Maurício Gruppi, Benjamin~D. Horne, and Sibel Adalı. 2021.
\newblock \href {http://arxiv.org/abs/2102.04567} {Nela-gt-2020: A large
  multi-labelled news dataset for the study of misinformation in news
  articles}.

\bibitem[{Gruppi et~al.(2023)Gruppi, Horne, and Adalı}]{gruppi2023nelagt2022}
Maurício Gruppi, Benjamin~D. Horne, and Sibel Adalı. 2023.
\newblock \href {http://arxiv.org/abs/2203.05659} {Nela-gt-2022: A large
  multi-labelled news dataset for the study of misinformation in news
  articles}.

\bibitem[{Guan et~al.(2023)Guan, Dodge, Wadden, Huang, and
  Peng}]{guan2023language}
Jian Guan, Jesse Dodge, David Wadden, Minlie Huang, and Hao Peng. 2023.
\newblock Language models hallucinate, but may excel at fact verification.
\newblock \emph{arXiv}.

\bibitem[{Hsu et~al.(2023)Hsu, Dai, Xiong, and Ku}]{Hsu2023IsET}
Yi-Li Hsu, Shih-Chieh Dai, Aiping Xiong, and Lun-Wei Ku. 2023.
\newblock \href {https://api.semanticscholar.org/CorpusID:264555229} {Is
  explanation the cure? misinformation mitigation in the short term and long
  term}.
\newblock \emph{EMNLP Findings}.

\bibitem[{Khashabi et~al.(2020)Khashabi, Min, Khot, Sabharwal, Tafjord, Clark,
  and Hajishirzi}]{khashabi-etal-2020-unifiedqa}
Daniel Khashabi, Sewon Min, Tushar Khot, Ashish Sabharwal, Oyvind Tafjord,
  Peter Clark, and Hannaneh Hajishirzi. 2020.
\newblock \href {https://doi.org/10.18653/v1/2020.findings-emnlp.171}
  {{UNIFIEDQA}: Crossing format boundaries with a single {QA} system}.
\newblock In \emph{Findings of the Association for Computational Linguistics:
  EMNLP 2020}, pages 1896--1907, Online. Association for Computational
  Linguistics.

\bibitem[{Kiela et~al.(2020)Kiela, Firooz, Mohan, Goswami, Singh, Ringshia, and
  Testuggine}]{Kiela2020TheHM}
Douwe Kiela, Hamed Firooz, Aravind Mohan, Vedanuj Goswami, Amanpreet Singh,
  Pratik Ringshia, and Davide Testuggine. 2020.
\newblock The hateful memes challenge: Detecting hate speech in multimodal
  memes.
\newblock \emph{NeurIPS}.

\bibitem[{Lee et~al.(2021)Lee, Li, Wang, Fung, Ma, tau Yih, and
  Khabsa}]{Lee2021OnUM}
Nayeon Lee, Belinda~Z. Li, Sinong Wang, Pascale Fung, Hao Ma, Wen tau Yih, and
  Madian Khabsa. 2021.
\newblock On unifying misinformation detection.
\newblock In \emph{North American Chapter of the Association for Computational
  Linguistics}.

\bibitem[{Liu et~al.(2024)Liu, Li, Li, Li, Zhang, Shen, and
  Lee}]{liu2024llavanext}
Haotian Liu, Chunyuan Li, Yuheng Li, Bo~Li, Yuanhan Zhang, Sheng Shen, and
  Yong~Jae Lee. 2024.
\newblock \href {https://llava-vl.github.io/blog/2024-01-30-llava-next/}
  {Llava-next: Improved reasoning, ocr, and world knowledge}.

\bibitem[{Morrish(2023)}]{wired-automated-factcheck}
Lydia Morrish. 2023.
\newblock \href
  {https://www.wired.com/story/fact-checkers-ai-chatgpt-misinformation/}
  {Fact-checkers are scrambling to fight disinformation with ai}.
\newblock \emph{Wired}.

\bibitem[{Nakamura et~al.(2020)Nakamura, Levy, and
  Wang}]{nakamura-etal-2020-fakeddit}
Kai Nakamura, Sharon Levy, and William~Yang Wang. 2020.
\newblock \href {https://aclanthology.org/2020.lrec-1.755} {{F}akeddit: A new
  multimodal benchmark dataset for fine-grained fake news detection}.
\newblock In \emph{Proceedings of the Twelfth Language Resources and Evaluation
  Conference}, pages 6149--6157, Marseille, France. European Language Resources
  Association.

\bibitem[{Nakov et~al.(2021)Nakov, Corney, Hasanain, Alam, Elsayed,
  Barr'on-Cedeno, Papotti, Shaar, and Martino}]{Nakov2021AutomatedFF}
Preslav Nakov, David Corney, Maram Hasanain, Firoj Alam, Tamer Elsayed, Alberto
  Barr'on-Cedeno, Paolo Papotti, Shaden Shaar, and Giovanni Da~San Martino.
  2021.
\newblock Automated fact-checking for assisting human fact-checkers.
\newblock In \emph{International Joint Conference on Artificial Intelligence}.

\bibitem[{Newell(1973)}]{Newell1973HumanPS}
Allen Newell. 1973.
\newblock Human problem solving.

\bibitem[{OpenAI(2022)}]{GPT35}
OpenAI. 2022.
\newblock Gpt-3.5.

\bibitem[{OpenAI(2023)}]{Achiam2023GPT4TR}
OpenAI. 2023.
\newblock \href {https://api.semanticscholar.org/CorpusID:257532815} {Gpt-4
  technical report}.

\bibitem[{P{\'e}rez-Rosas et~al.(2017)P{\'e}rez-Rosas, Kleinberg, Lefevre, and
  Mihalcea}]{PrezRosas2017AutomaticDO}
Ver{\'o}nica P{\'e}rez-Rosas, Bennett Kleinberg, Alexandra Lefevre, and Rada
  Mihalcea. 2017.
\newblock Automatic detection of fake news.
\newblock In \emph{International Conference on Computational Linguistics}.

\bibitem[{Radford et~al.(2021)Radford, Kim, Hallacy, Ramesh, Goh, Agarwal,
  Sastry, Askell, Mishkin, Clark, Krueger, and
  Sutskever}]{Radford2021LearningTV}
Alec Radford, Jong~Wook Kim, Chris Hallacy, Aditya Ramesh, Gabriel Goh,
  Sandhini Agarwal, Girish Sastry, Amanda Askell, Pamela Mishkin, Jack Clark,
  Gretchen Krueger, and Ilya Sutskever. 2021.
\newblock Learning transferable visual models from natural language
  supervision.
\newblock In \emph{International Conference on Machine Learning}.

\bibitem[{Rashkin et~al.(2017)Rashkin, Choi, Jang, Volkova, and
  Choi}]{rashkin-etal-2017-truth}
Hannah Rashkin, Eunsol Choi, Jin~Yea Jang, Svitlana Volkova, and Yejin Choi.
  2017.
\newblock \href {https://doi.org/10.18653/v1/D17-1317} {Truth of varying
  shades: Analyzing language in fake news and political fact-checking}.
\newblock In \emph{Proceedings of the 2017 Conference on Empirical Methods in
  Natural Language Processing}, pages 2931--2937, Copenhagen, Denmark.
  Association for Computational Linguistics.

\bibitem[{Schuster et~al.(2021)Schuster, Fisch, and
  Barzilay}]{schuster-etal-2021-get}
Tal Schuster, Adam Fisch, and Regina Barzilay. 2021.
\newblock \href {https://doi.org/10.18653/v1/2021.naacl-main.52} {Get your
  vitamin {C}! robust fact verification with contrastive evidence}.
\newblock In \emph{Proceedings of the 2021 Conference of the North American
  Chapter of the Association for Computational Linguistics: Human Language
  Technologies}, pages 624--643, Online. Association for Computational
  Linguistics.

\bibitem[{Shimodaira(2000)}]{SHIMODAIRA2000227}
Hidetoshi Shimodaira. 2000.
\newblock \href {https://doi.org/https://doi.org/10.1016/S0378-3758(00)00115-4}
  {Improving predictive inference under covariate shift by weighting the
  log-likelihood function}.
\newblock \emph{Journal of Statistical Planning and Inference}, 90(2):227--244.

\bibitem[{Wan et~al.(2024)Wan, Feng, Tan, Wang, Tsvetkov, and
  Luo}]{Wan2024DELLGR}
Herun Wan, Shangbin Feng, Zhaoxuan Tan, Heng Wang, Yulia Tsvetkov, and Minnan
  Luo. 2024.
\newblock \href {https://api.semanticscholar.org/CorpusID:267740574} {Dell:
  Generating reactions and explanations for llm-based misinformation
  detection}.
\newblock \emph{ArXiv}, abs/2402.10426.

\bibitem[{Wang(2017)}]{Wang2017LiarLP}
William~Yang Wang. 2017.
\newblock “liar, liar pants on fire”: A new benchmark dataset for fake news
  detection.
\newblock In \emph{Annual Meeting of the Association for Computational
  Linguistics}.

\bibitem[{Yang et~al.(2019)Yang, Shu, Wang, Gu, Wu, and
  Liu}]{Yang2019UnsupervisedFN}
Shuo Yang, Kai Shu, Suhang Wang, Renjie Gu, Fan Wu, and Huan Liu. 2019.
\newblock Unsupervised fake news detection on social media: A generative
  approach.
\newblock In \emph{AAAI Conference on Artificial Intelligence}.

\bibitem[{Yao et~al.(2022)Yao, Shah, Sun, Cho, and Huang}]{Yao2022EndtoEndMF}
Barry~Menglong Yao, Aditya Shah, Lichao Sun, Jin-Hee Cho, and Lifu Huang. 2022.
\newblock End-to-end multimodal fact-checking and explanation generation: A
  challenging dataset and models.
\newblock \emph{ArXiv}, abs/2205.12487.

\bibitem[{Zellers et~al.(2019)Zellers, Holtzman, Rashkin, Bisk, Farhadi,
  Roesner, and Choi}]{zellers2019neuralfakenews}
Rowan Zellers, Ari Holtzman, Hannah Rashkin, Yonatan Bisk, Ali Farhadi,
  Franziska Roesner, and Yejin Choi. 2019.
\newblock \href
  {http://papers.nips.cc/paper/9106-defending-against-neural-fake-news.pdf}
  {Defending against neural fake news}.
\newblock In H.~Wallach, H.~Larochelle, A.~Beygelzimer, F.~d\textquotesingle
  Alch\'{e}-Buc, E.~Fox, and R.~Garnett, editors, \emph{Advances in Neural
  Information Processing Systems 32}, pages 9054--9065. Curran Associates, Inc.

\end{thebibliography}
\bibliographystyle{acl_natbib}

\clearpage

\appendix

\section{Appendix}
\label{sec:appendix}
\subsection{ChatGPT Usage}

ChatGPT has been used for writing simple scripts, including normalizing datasets into common schemas or merging different tsv files into a single file.

\subsection{Human Evaluation Details}

\begin{figure*}[t]
    \centering
    \includegraphics[width=.9\textwidth]{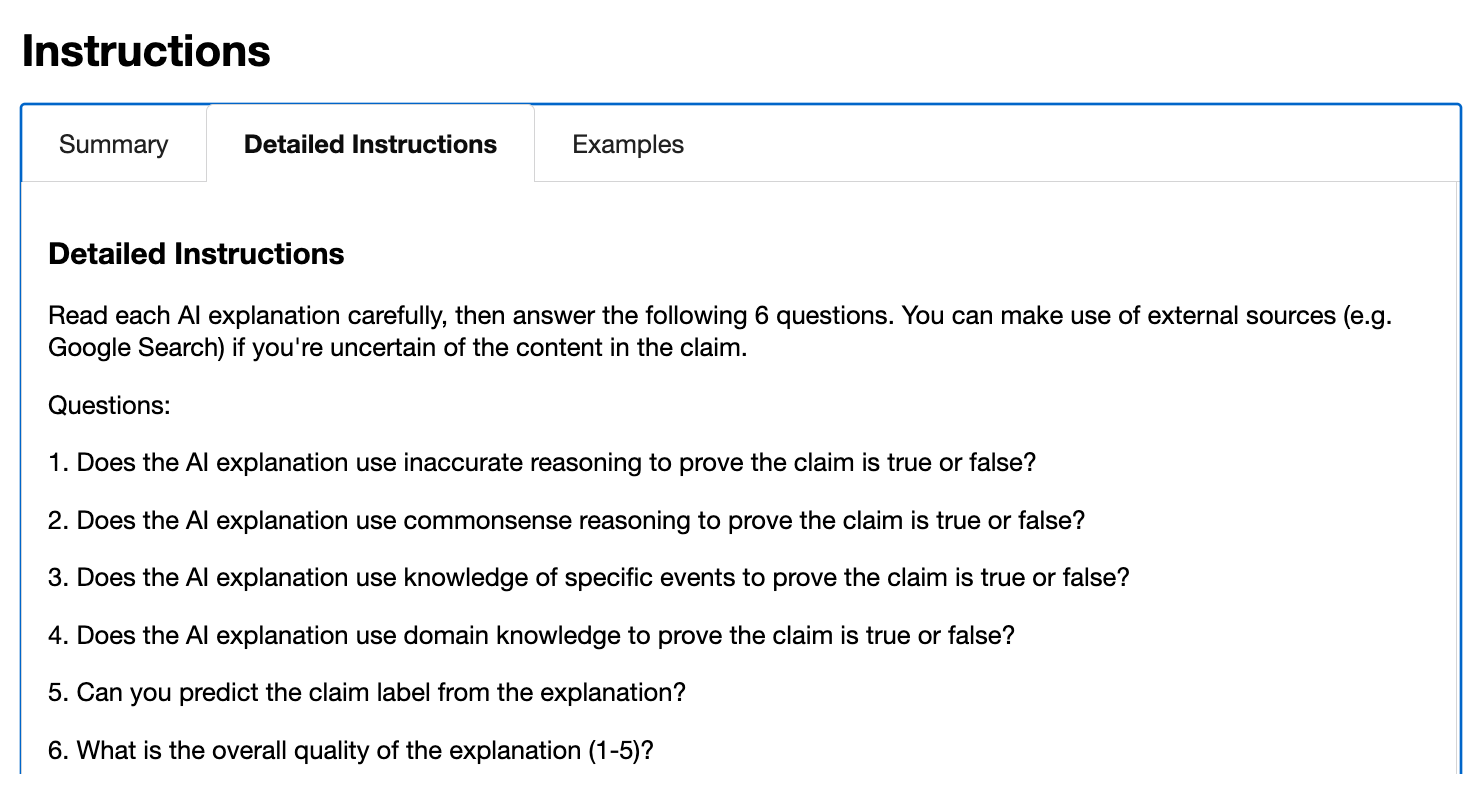}
    \caption{Full Instructions for Amazon Mechanical Turk Task.}
    \label{fig:appendix1}
\end{figure*}

\begin{figure*}[t]
    \centering
    \includegraphics[width=.9\textwidth]{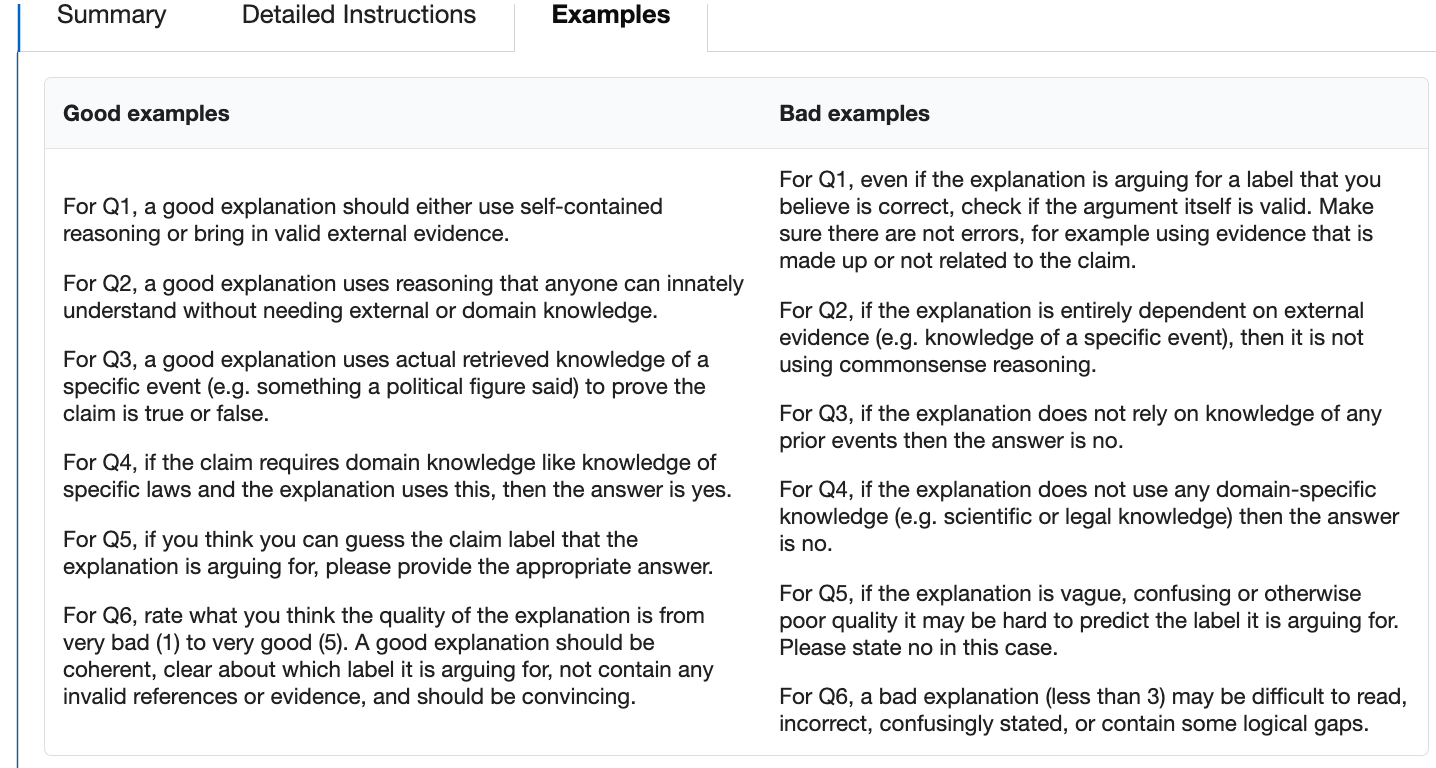}
    \caption{Explanations for Answering Amazon Mechanical Turk Questions.}
    \label{fig:appendix2}
\end{figure*}

Annotators were paid \$0.20 per explanation evaluation, which we judged to be a fair wage based on estimated time required to complete the task. Full instructions given to annotators are shown in Figures \ref{fig:appendix1} and \ref{fig:appendix2}. We obtained the appropriate IRB exemption approval for the study. 

\subsection{Supplementary Results}
\renewcommand{\arraystretch}{1.2} 
\begin{table*}[!ht]
    \centering
    \resizebox{\textwidth}{!}{ 
    \begin{tabular}{|l|c|c|c|c|c|c|c|c|c|c|c|c|}
        \textbf{Dataset Mixture} & \textbf{MC} & \textbf{FK} & \textbf{PH} & \textbf{Pre-V} & \textbf{Post-V} & \textbf{U-R} & \textbf{MRF} & \textbf{HX} & \textbf{MMHS} & \textbf{TX} & \textbf{HM} & \textbf{PS} \\ \hline
        FK & 54.87\ & 89.31\ & 43.36\ & 52.70\ & 44.80\ & 53.50\ & 57.77\ & 54.89\ & 65.77\ & 49.15\ & 50.00\ & 47.98\ \\ 
        FK + PH & 39.23\ & 88.73\ & 76.16\ & 43.30\ & 44.30\ & 44.60\ & 60.86\ & 63.77\ & 63.80\ & 47.55\ & 50.00\ & 51.65\ \\
        FV & 38.25\ & 60.76\ & 60.66\ & 49.40\ & 48.00\ & 50.40\ & 47.92\ & 38.62\ & 52.70\ & 54.04\ & 48.40\ & 50.12\ \\ 
        FV + FK & 36.69\ & 87.99\ & 59.01\ & 49.80\ & 50.30\ & 48.70\ & 41.92\ & 38.72\ & 63.44\ & 52.55\ & 50.40\ & 49.24\ \\ 
        FV + FK + PH & 38.78\ & 88.07\ & 74.61\ & 48.40\ & 49.90\ & 49.90\ & 39.01\ & 28.33\ & 62.30\ & 57.77\ & 52.80\ & 47.47\ \\ 
        FV + NGT & 32.88\ & 60.33\ & 52.23\ & 35.30\ & 24.20\ & 17.20\ & 74.16\ & 70.32\ & 74.06\ & 53.62\ & 50.20\ & 51.97\ \\ 
        FV + NGT + FK & 36.16\ & 90.72\ & 63.08\ & 41.90\ & 32.60\ & 23.70\ & 69.82\ & 72.45\ & 69.19\ & 50.11\ & 51.60\ & 52.80\ \\
        FV + NGT + MC & 46.03\ & 60.41\ & 56.59\ & 35.40\ & 25.90\ & 18.70\ & 78.55\ & 72.25\ & 39.28\ & 50.21\ & 50.00\ & 52.20\ \\
        FV + NGT + PH & 32.27\ & 60.37\ & 72.19\ & 32.70\ & 24.40\ & 16.90\ & 75.19\ & 68.24\ & 74.02\ & 54.79\ & 50.00\ & 53.41\ \\
        FV + NGT + VC & 29.36\ & 60.37\ & 34.01\ & 32.80\ & 25.60\ & 16.10\ & 74.47\ & 70.06\ & 74.01\ & 52.23\ & 50.00\ & 52.25\ \\
        FV + PH & 37.80\ & 59.29\ & 74.61\ & 60.30\ & 59.20\ & 60.00\ & 44.16\ & 32.07\ & 71.92\ & 51.81\ & 50.20\ & 42.10\ \\
        FV + VC & 35.50\ & 39.70\ & 48.64\ & 50.00\ & 50.40\ & 49.80\ & 37.62\ & 26.20\ & 25.89\ & 55.53\ & 50.00\ & 47.57\ \\
        FV + VC + FK & 35.42\ & 89.11\ & 37.98\ & 49.30\ & 45.30\ & 51.30\ & 49.57\ & 48.49\ & 66.31\ & 51.38\ & 51.20\ & 51.09\ \\
        FV + VC + PH & 39.03\ & 39.45\ & 64.92\ & 50.10\ & 50.30\ & 50.10\ & 35.20\ & 26.35\ & 26.50\ & 55.85\ & 49.60\ & 47.43\ \\
        HX & 57.61\ & 60.46\ & 42.25\ & 48.50\ & 47.60\ & 46.40\ & 65.56\ & 80.67\ & 43.83\ & 54.89\ & 48.80\ & 53.92\ \\ 
        MC & 49.18\ & 39.66\ & 35.95\ & 50.00\ & 50.00\ & 50.00\ & 37.75\ & 26.20\ & 25.88\ & 56.28\ & 50.00\ & 47.84\ \\ 
        MC + FK & 47.87\ & 87.34\ & 32.66\ & 44.70\ & 45.60\ & 46.50\ & 57.55\ & 69.39\ & 58.01\ & 55.96\ & 51.00\ & 52.43\ \\
        MC + FK + PH & 49.63\ & 89.22\ & 73.06\ & 48.10\ & 50.40\ & 51.60\ & 59.70\ & 68.04\ & 64.61\ & 49.04\ & 51.40\ & 50.72\ \\
        MC + FV & 45.13\ & 62.18\ & 47.58\ & 47.40\ & 50.60\ & 49.00\ & 48.32\ & 41.53\ & 48.82\ & 58.09\ & 48.20\ & 45.34\ \\
        MC + FV + FK & 46.56\ & 89.01\ & 47.97\ & 49.30\ & 49.80\ & 49.60\ & 38.69\ & 28.90\ & 64.18\ & 56.38\ & 51.40\ & 48.08\ \\
        MC + FV + PH & 48.81\ & 39.71\ & 76.55\ & 50.10\ & 50.10\ & 50.10\ & 38.47\ & 26.51\ & 25.88\ & 56.17\ & 50.00\ & 47.84\ \\
        MC + FV + VC & 46.23\ & 40.30\ & 40.99\ & 49.90\ & 49.90\ & 50.00\ & 37.80\ & 26.51\ & 25.94\ & 56.17\ & 49.80\ & 47.94\ \\
        MC + HX & 49.06\ & 40.53\ & 41.18\ & 40.00\ & 44.90\ & 44.60\ & 64.53\ & 80.15\ & 26.84\ & 58.30\ & 49.60\ & 59.99\ \\ 
        MC + PH & 47.50\ & 39.69\ & 76.16\ & 50.00\ & 50.00\ & 50.00\ & 37.75\ & 26.20\ & 25.98\ & 56.28\ & 49.60\ & 47.84\ \\ 
        MC + TX & 49.84\ & 40.38\ & 32.56\ & 44.90\ & 51.20\ & 44.80\ & 64.89\ & 73.96\ & 30.83\ & 50.43\ & 49.20\ & 53.36\ \\ 
        MC + VC & 45.74\ & 48.48\ & 43.02\ & 49.50\ & 48.80\ & 49.50\ & 41.78\ & 28.69\ & 33.52\ & 54.68\ & 51.00\ & 47.57\ \\ 
        MC + VC  + FK & 43.00\ & 89.45\ & 36.34\ & 48.50\ & 49.30\ & 52.70\ & 38.42\ & 31.86\ & 67.58\ & 56.60\ & 51.00\ & 48.22\ \\
        MC + VC  + PH & 45.41\ & 46.22\ & 68.02\ & 50.00\ & 49.30\ & 50.00\ & 37.89\ & 26.40\ & 33.57\ & 56.49\ & 51.80\ & 47.66\ \\ 
        PH & 37.88\ & 55.78\ & 76.36\ & 50.00\ & 50.10\ & 49.00\ & 62.11\ & 73.39\ & 48.60\ & 43.83\ & 48.80\ & 52.11\ \\ 
        TX & 48.90\ & 60.54\ & 52.79\ & 49.30\ & 48.90\ & 48.70\ & 39.72\ & 57.90\ & 62.56\ & 70.43\ & 49.00\ & 53.41\ \\ 
        VC & 35.38\ & 57.84\ & 41.09\ & 50.60\ & 51.00\ & 50.40\ & 37.04\ & 28.38\ & 36.70\ & 56.17\ & 50.60\ & 47.24\ \\ 
        VC + FK & 35.22\ & 89.40\ & 39.83\ & 46.30\ & 51.70\ & 50.20\ & 46.04\ & 38.88\ & 66.61\ & 50.32\ & 51.40\ & 45.11\ \\ 
        VC + NGT + FK & 33.33\ & 89.22\ & 31.01\ & 41.40\ & 31.80\ & 23.30\ & 56.92\ & 52.60\ & 62.82\ & 48.83\ & 50.00\ & 51.69\ \\ 
        VC + NGT + MC & 44.55\ & 62.78\ & 38.37\ & 34.70\ & 25.20\ & 19.50\ & 71.20\ & 65.85\ & 64.11\ & 52.77\ & 52.40\ & 53.13\ \\ 
        VC + NGT + PH & 36.69\ & 62.17\ & 70.45\ & 31.90\ & 25.60\ & 19.80\ & 72.59\ & 65.28\ & 62.61\ & 54.04\ & 49.20\ & 53.36\ \\ 
        VC + PH & 34.73\ & 60.45\ & 65.79\ & 49.50\ & 49.10\ & 50.20\ & 37.98\ & 26.20\ & 73.93\ & 56.06\ & 49.80\ & 47.66\ \\ 
        VC + PH + FK & 38.37\ & 88.43\ & 70.45\ & 45.50\ & 51.30\ & 49.30\ & 58.49\ & 56.91\ & 60.92\ & 53.40\ & 51.20\ & 49.79\ \\ \hline
    \end{tabular}
    }
    \caption{ The results from the CLIP-based verifier are shown above. Abbreviations include: \textbf{FK} - Fakeddit, \textbf{FV} - Fever, \textbf{HM} - HatefulMemes, \textbf{HX} - HateXplain, \textbf{MC} - Mocheg, \textbf{MMHS} - MultiModal Hate Speech 150k, \textbf{MRF} - Misinformation Reaction Framework, \textbf{NGT} - NelaGT2022, \textbf{PH} - PubHealth, \textbf{PS} - PStance, \textbf{TX} - Toxigen, \textbf{VC} - VitaminC.}
    \label{tab:second}
\end{table*}

\renewcommand{\arraystretch}{1.2} 
\begin{table*}[!ht]
    \centering
    \resizebox{\textwidth}{!}{ 
    \begin{tabular}{|l|c|c|c|c|c|c|c|c|c|c|c|c|}
        \textbf{Dataset Mixtures} & \textbf{MC} & \textbf{FK} & \textbf{PH} & \textbf{Pre-V} & \textbf{Post-V} & \textbf{U-R} & \textbf{MRF} & \textbf{HX} & \textbf{MMHS} & \textbf{TX} & \textbf{HM} & \textbf{PS} \\ \hline 
        FK & 57.13\ & 91.41\ & 46.10\ & 50.90\ & 47.90\ & 49.40\ & 61.71\ & 72.71\ & 70.36\ & 44.36\ & 51.20\ & 51.65\ \\ 
        FK + PH & 39.35\ & 91.21\ & 72.48\ & 40.30\ & 42.50\ & 36.90\ & 59.43\ & 65.12\ & 64.37\ & 47.66\ & 52.00\ & 48.86\ \\ 
        FV & 38.41\ & 39.66\ & 58.91\ & 49.40\ & 52.90\ & 48.30\ & 47.83\ & 48.65\ & 25.88\ & 42.45\ & 50.00\ & 46.96\ \\ 
        FV + FK & 36.69\ & 91.37\ & 59.59\ & 49.50\ & 49.60\ & 49.80\ & 38.83\ & 27.44\ & 65.82\ & 55.64\ & 50.40\ & 48.40\ \\ 
        FV + FK + PH & 38.17\ & 92.27\ & 75.39\ & 49.10\ & 49.50\ & 49.70\ & 38.83\ & 30.04\ & 70.92\ & 55.74\ & 50.20\ & 48.22\ \\ 
        FV + NGT & 37.26\ & 39.41\ & 57.56\ & 31.00\ & 22.40\ & 16.10\ & 76.18\ & 69.70\ & 26.99\ & 52.02\ & 49.80\ & 52.53\ \\ 
        FV + NGT + FK & 36.45\ & 92.32\ & 62.50\ & 32.50\ & 26.60\ & 17.80\ & 79.98\ & 72.61\ & 70.18\ & 53.19\ & 50.80\ & 52.43\ \\ 
        FV + NGT + MC & 48.28\ & 58.93\ & 60.95\ & 28.30\ & 22.30\ & 17.60\ & 76.62\ & 68.50\ & 48.23\ & 53.94\ & 50.60\ & 52.39\ \\ 
        FV + NGT + PH & 40.50\ & 39.68\ & 76.65\ & 31.40\ & 22.40\ & 16.00\ & 77.03\ & 69.70\ & 25.88\ & 51.60\ & 49.80\ & 51.55\ \\ 
        FV + NGT + VC & 35.14\ & 52.80\ & 36.82\ & 31.10\ & 23.90\ & 18.10\ & 74.07\ & 71.83\ & 40.54\ & 48.40\ & 51.20\ & 51.32\ \\ 
        FV + PH & 40.66\ & 39.66\ & 74.42\ & 50.30\ & 50.00\ & 50.00\ & 62.16\ & 73.80\ & 25.88\ & 43.94\ & 50.00\ & 52.16\ \\ 
        FV + VC & 34.93\ & 39.66\ & 32.85\ & 51.70\ & 50.80\ & 49.20\ & 42.68\ & 42.67\ & 25.92\ & 52.45\ & 50.00\ & 44.51\ \\ 
        FV + VC + FK & 35.71\ & 91.81\ & 40.21\ & 54.30\ & 48.20\ & 52.00\ & 37.84\ & 28.59\ & 67.61\ & 56.38\ & 51.80\ & 47.52\ \\ 
        FV + VC + PH & 39.84\ & 39.66\ & 67.64\ & 50.10\ & 50.00\ & 49.60\ & 40.62\ & 30.15\ & 25.88\ & 53.94\ & 50.00\ & 47.61\ \\ 
        HX & 57.43\ & 39.66\ & 50.96\ & 44.90\ & 46.50\ & 43.80\ & 49.75\ & 81.44\ & 25.88\ & 65.11\ & 50.00\ & 55.49\ \\ 
        MC & 50.49\ & 48.52\ & 41.96\ & 50.00\ & 50.00\ & 50.00\ & 62.25\ & 73.80\ & 44.73\ & 43.72\ & 48.60\ & 52.16\ \\ 
        MC + FK & 48.16\ & 91.44\ & 36.92\ & 49.50\ & 49.20\ & 49.10\ & 38.11\ & 27.81\ & 64.63\ & 56.38\ & 50.60\ & 47.89\ \\ 
        MC + FK + PH & 50.08\ & 92.46\ & 73.84\ & 49.50\ & 49.60\ & 49.90\ & 38.47\ & 35.86\ & 71.82\ & 56.60\ & 50.80\ & 50.02\ \\ 
        MC + FV & 49.02\ & 43.25\ & 42.44\ & 50.00\ & 48.60\ & 49.90\ & 42.59\ & 28.79\ & 29.68\ & 56.91\ & 53.20\ & 48.31\ \\ 
        MC + FV + FK & 47.05\ & 92.10\ & 53.68\ & 50.00\ & 50.00\ & 49.90\ & 37.80\ & 26.25\ & 69.55\ & 56.28\ & 51.20\ & 47.89\ \\ 
        MC + FV + PH & 50.86\ & 61.18\ & 74.81\ & 50.90\ & 49.60\ & 51.00\ & 46.62\ & 31.86\ & 51.11\ & 55.43\ & 49.80\ & 47.52\ \\ 
        MC + FV + VC & 42.30\ & 49.27\ & 43.31\ & 51.20\ & 50.20\ & 50.60\ & 50.56\ & 43.97\ & 33.40\ & 52.55\ & 49.80\ & 48.86\ \\ 
        MC + HX & 50.04\ & 55.27\ & 35.37\ & 49.30\ & 48.30\ & 49.10\ & 68.25\ & 75.94\ & 59.49\ & 48.30\ & 51.00\ & 53.41\ \\ 
        MC + PH & 51.68\ & 43.53\ & 75.97\ & 50.00\ & 50.00\ & 50.10\ & 37.80\ & 26.40\ & 28.22\ & 56.17\ & 50.60\ & 47.57\ \\ 
        MC + TX & 49.26\ & 39.87\ & 32.66\ & 50.00\ & 48.00\ & 48.90\ & 64.44\ & 73.80\ & 26.50\ & 51.49\ & 51.60\ & 52.76\ \\ 
        MC + VC & 43.61\ & 46.90\ & 53.20\ & 50.60\ & 45.40\ & 49.60\ & 47.56\ & 60.50\ & 47.16\ & 56.60\ & 52.00\ & 52.11\ \\ 
        MC + VC + FK & 47.75\ & 91.94\ & 41.96\ & 43.00\ & 43.90\ & 47.50\ & 63.14\ & 62.73\ & 68.07\ & 54.36\ & 51.00\ & 52.39\ \\ 
        MC + VC + PH & 46.52\ & 43.66\ & 70.93\ & 48.50\ & 47.80\ & 49.90\ & 39.54\ & 33.16\ & 39.01\ & 54.79\ & 52.60\ & 48.77\ \\ 
        PH & 41.40\ & 39.66\ & 76.45\ & 50.70\ & 46.60\ & 52.70\ & 66.32\ & 68.04\ & 25.88\ & 53.94\ & 50.00\ & 47.66\ \\ 
        TX & 59.99\ & 39.66\ & 46.71\ & 49.10\ & 49.70\ & 50.60\ & 62.61\ & 73.91\ & 25.88\ & 46.60\ & 50.00\ & 52.85\ \\ 
        VC & 35.79\ & 39.66\ & 36.82\ & 50.30\ & 51.50\ & 51.50\ & 44.74\ & 41.74\ & 25.87\ & 52.34\ & 50.00\ & 48.77\ \\ 
        VC + FK & 33.70\ & 92.27\ & 35.47\ & 49.00\ & 50.00\ & 49.50\ & 38.78\ & 28.27\ & 70.61\ & 56.49\ & 51.20\ & 49.61\ \\ 
        VC + NGT + FK & 34.81\ & 91.95\ & 37.89\ & 34.90\ & 27.50\ & 22.70\ & 68.61\ & 63.36\ & 69.35\ & 49.68\ & 50.60\ & 51.60\ \\ 
        VC + NGT + MC & 46.52\ & 45.32\ & 42.34\ & 33.30\ & 28.20\ & 18.30\ & 74.88\ & 66.53\ & 39.40\ & 51.91\ & 52.40\ & 53.27\ \\ 
        VC + NGT + PH & 37.76\ & 39.65\ & 68.22\ & 32.30\ & 24.50\ & 17.70\ & 73.26\ & 67.62\ & 25.88\ & 54.36\ & 50.00\ & 53.08\ \\ 
        VC + PH & 37.84\ & 39.88\ & 69.19\ & 48.60\ & 50.10\ & 49.60\ & 43.48\ & 32.90\ & 28.72\ & 55.43\ & 49.00\ & 47.61\ \\ 
        VC + PH + FK & 36.57\ & 91.13\ & 71.80\ & 43.70\ & 45.10\ & 46.50\ & 52.40\ & 35.65\ & 71.64\ & 56.91\ & 51.00\ & 50.16\ \\ \hline 
    \end{tabular}
    }
    \caption{ The results from the CLIP-large verifier are shown above. Abbreviations are same as Table \ref{tab:second}.}
\end{table*}

\renewcommand{\arraystretch}{1.2} 
\begin{table*}[!ht]
    \centering
    \resizebox{\textwidth}{!}{ 
    \begin{tabular}{|l|c|c|c|c|c|c|c|c|c|c|c|c|}
        \textbf{Dataset Mixtures} & \textbf{MC} & \textbf{FK} & \textbf{PH} & \textbf{Pre-V} & \textbf{Post-V} & \textbf{U-R} & \textbf{MRF} & \textbf{HX} & \textbf{MMHS} & \textbf{TX} & \textbf{HM} & \textbf{PS} \\ \hline 
        FK & 58.10\ & 91.87\ & 49.75\ & 46.20\ & 47.80\ & 43.80\ & 62.38\ & 58.58\ & 63.78\ & 47.34\ & 50.80\ & 50.86\ \\ 
        FK + PH & 39.84\ & 92.96\ & 75.97\ & 45.50\ & 46.50\ & 47.50\ & 43.62\ & 58.26\ & 70.90\ & 60.43\ & 51.00\ & 51.23\ \\ 
        FV & 37.59\ & 39.66\ & 58.62\ & 48.00\ & 52.70\ & 46.40\ & 42.32\ & 48.18\ & 25.88\ & 46.49\ & 50.00\ & 45.57\ \\ 
        FV + FK & 37.26\ & 92.63\ & 57.36\ & 50.30\ & 49.00\ & 50.10\ & 38.42\ & 27.18\ & 71.95\ & 56.17\ & 50.40\ & 48.35\ \\ 
        FV + FK + PH & 39.93\ & 91.62\ & 74.22\ & 50.00\ & 50.00\ & 50.10\ & 37.89\ & 26.20\ & 70.13\ & 56.28\ & 51.00\ & 47.80\ \\ 
        FV + NGT & 37.71\ & 39.89\ & 55.23\ & 33.80\ & 23.90\ & 17.20\ & 69.28\ & 63.20\ & 25.99\ & 53.09\ & 50.00\ & 52.85\ \\ 
        FV + NGT + FK & 34.68\ & 92.34\ & 61.05\ & 31.30\ & 23.80\ & 17.70\ & 79.85\ & 72.40\ & 72.19\ & 51.17\ & 50.00\ & 53.41\ \\ 
        FV + NGT + MC & 48.85\ & 44.45\ & 56.69\ & 31.70\ & 22.10\ & 15.40\ & 75.91\ & 69.49\ & 31.95\ & 53.40\ & 50.60\ & 52.20\ \\ 
        FV + NGT + PH & 40.21\ & 39.66\ & 74.13\ & 32.00\ & 24.80\ & 17.90\ & 73.58\ & 61.49\ & 25.88\ & 54.15\ & 50.00\ & 52.76\ \\ 
        FV + NGT + VC & 35.54\ & 44.78\ & 33.53\ & 31.00\ & 27.50\ & 15.10\ & 74.12\ & 67.31\ & 30.46\ & 51.17\ & 50.80\ & 53.41\ \\ 
        FV + PH & 39.56\ & 40.35\ & 75.00\ & 48.40\ & 53.00\ & 50.00\ & 40.53\ & 26.77\ & 25.94\ & 55.32\ & 50.00\ & 48.49\ \\ 
        FV + VC & 36.73\ & 39.66\ & 39.63\ & 51.10\ & 50.10\ & 49.90\ & 41.42\ & 33.58\ & 25.89\ & 56.06\ & 50.00\ & 47.15\ \\ 
        FV + VC + FK & 33.58\ & 92.22\ & 32.17\ & 51.50\ & 47.80\ & 45.80\ & 41.24\ & 45.63\ & 69.84\ & 53.83\ & 50.80\ & 49.42\ \\ 
        FV + VC + PH & 39.80\ & 39.66\ & 70.25\ & 50.50\ & 50.00\ & 50.50\ & 39.50\ & 29.11\ & 25.88\ & 55.32\ & 50.00\ & 47.20\ \\ 
        HX & 59.44\ & 39.66\ & 43.26\ & 48.10\ & 48.10\ & 49.40\ & 38.42\ & 72.35\ & 25.88\ & 64.04\ & 50.00\ & 49.10\ \\ 
        MC & 50.61\ & 39.78\ & 27.33\ & 49.40\ & 49.70\ & 50.00\ & 38.42\ & 26.35\ & 25.86\ & 55.74\ & 49.80\ & 48.08\ \\ 
        MC + FK & 50.82\ & 92.80\ & 38.18\ & 49.90\ & 50.20\ & 51.40\ & 59.34\ & 68.81\ & 69.78\ & 48.19\ & 50.20\ & 51.97\ \\ 
        MC + FK + PH & 47.83\ & 91.66\ & 74.03\ & 50.50\ & 51.40\ & 50.30\ & 38.51\ & 38.46\ & 68.78\ & 57.02\ & 51.40\ & 47.94\ \\ 
        MC + FV & 49.47\ & 49.47\ & 46.71\ & 50.40\ & 50.30\ & 49.70\ & 42.10\ & 28.17\ & 30.28\ & 55.96\ & 50.40\ & 47.94\ \\ 
        MC + FV + FK & 49.18\ & 92.49\ & 50.19\ & 51.20\ & 50.00\ & 50.40\ & 41.02\ & 37.84\ & 69.54\ & 57.55\ & 49.40\ & 52.25\ \\ 
        MC + FV + PH & 48.69\ & 58.42\ & 72.19\ & 49.10\ & 48.60\ & 49.50\ & 42.01\ & 36.49\ & 50.58\ & 54.89\ & 52.00\ & 48.08\ \\ 
        MC + FV + VC & 46.11\ & 48.21\ & 51.55\ & 50.30\ & 49.70\ & 50.60\ & 37.89\ & 26.92\ & 40.99\ & 55.11\ & 50.40\ & 47.71\ \\ 
        MC + HX & 50.45\ & 45.31\ & 40.02\ & 43.90\ & 45.40\ & 43.90\ & 45.32\ & 82.22\ & 34.60\ & 70.43\ & 54.20\ & 53.55\ \\ 
        MC + PH & 51.23\ & 39.71\ & 75.68\ & 50.00\ & 50.10\ & 50.00\ & 37.62\ & 26.46\ & 25.90\ & 56.06\ & 50.00\ & 48.08\ \\ 
        MC + TX & 50.20\ & 39.68\ & 26.94\ & 49.90\ & 48.90\ & 48.70\ & 64.17\ & 73.86\ & 26.06\ & 48.72\ & 50.00\ & 52.43\ \\ 
        MC + VC & 41.40\ & 47.31\ & 44.48\ & 51.90\ & 50.80\ & 48.20\ & 42.45\ & 45.69\ & 49.21\ & 55.96\ & 54.40\ & 48.45\ \\ 
        MC + VC  + FK & 47.46\ & 93.42\ & 44.57\ & 51.00\ & 49.60\ & 51.30\ & 37.35\ & 33.89\ & 71.73\ & 54.57\ & 50.60\ & 47.24\ \\ 
        MC + VC  + PH & 45.37\ & 41.18\ & 67.44\ & 48.60\ & 49.10\ & 50.10\ & 38.56\ & 32.48\ & 37.28\ & 55.53\ & 51.80\ & 49.42\ \\ 
        PH & 41.52\ & 39.66\ & 77.81\ & 50.20\ & 50.20\ & 50.10\ & 66.23\ & 72.40\ & 25.88\ & 47.98\ & 50.00\ & 50.86\ \\ 
        TX & 60.05\ & 39.66\ & 43.26\ & 49.10\ & 49.60\ & 50.20\ & 62.70\ & 73.80\ & 25.88\ & 45.85\ & 50.00\ & 52.57\ \\ 
        VC & 35.38\ & 39.43\ & 34.40\ & 51.00\ & 47.90\ & 48.30\ & 52.35\ & 65.44\ & 27.17\ & 45.74\ & 50.60\ & 51.65\ \\ 
        VC + FK & 32.15\ & 92.19\ & 45.25\ & 49.70\ & 49.20\ & 52.60\ & 33.05\ & 34.46\ & 71.72\ & 49.15\ & 50.60\ & 47.38\ \\ 
        VC + NGT + FK & 32.92\ & 92.67\ & 41.28\ & 33.70\ & 27.00\ & 22.50\ & 79.09\ & 74.12\ & 72.75\ & 50.11\ & 50.20\ & 51.88\ \\ 
        VC + NGT + MC & 44.88\ & 55.55\ & 49.61\ & 35.70\ & 25.00\ & 18.90\ & 68.20\ & 61.80\ & 52.28\ & 54.89\ & 55.20\ & 52.16\ \\ 
        VC + NGT + PH & 39.11\ & 39.67\ & 67.64\ & 29.40\ & 21.40\ & 16.50\ & 75.41\ & 63.62\ & 25.92\ & 54.68\ & 50.00\ & 52.90\ \\ 
        VC + PH & 39.39\ & 39.66\ & 70.25\ & 49.50\ & 47.30\ & 48.80\ & 49.71\ & 46.57\ & 25.88\ & 51.81\ & 50.00\ & 51.18\ \\ 
        VC + PH + FK & 39.15\ & 91.95\ & 68.51\ & 47.20\ & 47.10\ & 45.50\ & 45.45\ & 41.68\ & 69.15\ & 53.51\ & 49.20\ & 48.22\ \\ \hline
    \end{tabular}
    }
    \caption{ The results from the CLIP-large-336 verifier are shown above. Abbreviations are same as Table \ref{tab:second}.}
\end{table*}

\begin{table*}[!ht]
    \centering
    \begin{tabular}{|l|l|l|}
        Model & Method & F1-Score \\ \hline
        CLIP-base & Baseline & 49.18 \\ 
        CLIP-base & Random & 49.47 \\ 
        CLIP-base & Opposite & 53.56 \\ 
        CLIP-base & All & 48.40 \\ 
        CLIP-base & Always Supports & 48.16 \\ 
        CLIP-base & Always Refutes & 49.39 \\ 
        CLIP-base & Always NEI & 48.98 \\ 
        CLIP-base & Guided & 50.53 \\ 
        CLIP-base & Oracle & 60.69 \\ \hline
        CLIP-large & Baseline & 50.49 \\ 
        CLIP-large & Random & 50.66 \\ 
        CLIP-large & Opposite & 51.52 \\ 
        CLIP-large & All & 49.63 \\ 
        CLIP-large & Always Supports & 49.80 \\ 
        CLIP-large & Always Refutes & 50.45 \\ 
        CLIP-large & Always NEI & 52.21 \\ 
        CLIP-large & Guided & 50.90 \\ 
        CLIP-large & Oracle & 62.98 \\ \hline
        CLIP-large-336 & Baseline & 50.98 \\ 
        CLIP-large-336 & Random & 52.09\\ 
        CLIP-large-336 & Opposite & 51.72 \\ 
        CLIP-large-336 & All & 52.99 \\ 
        CLIP-large-336 & Always Supports & 50.66 \\ 
        CLIP-large-336 & Always Refutes & 52.33 \\ 
        CLIP-large-336 & Always NEI & 51.35 \\ 
        CLIP-large-336 & Guided & 51.64 \\ 
        CLIP-large-336 & Oracle & 62.69 \\ \hline
    \end{tabular}
    \caption{The Mocheg results with CLIP models with explanations generated by GPT-3.5-turbo. }
\end{table*}

\begin{table*}[!ht]
    \centering
    \begin{tabular}{|l|l|l|}
        Model & Method & F1-Score \\ \hline
        CLIP-base & Baseline & 49.18 \\ 
        CLIP-base & Random & 50.57 \\ 
        CLIP-base & Opposite & 49.34 \\ 
        CLIP-base & All & 50.57 \\ 
        CLIP-base & Always Supports & 51.97 \\ 
        CLIP-base & Always Refutes & 50.57 \\ 
        CLIP-base & Always NEI & 52.05 \\ 
        CLIP-base & Guided & 51.02 \\ 
        CLIP-base & Oracle & 59.21 \\ \hline
        CLIP-large & Baseline & 50.49 \\ 
        CLIP-large & Random & 50.94 \\ 
        CLIP-large & Opposite & 51.97 \\ 
        CLIP-large & All & 49.8 \\ 
        CLIP-large & Always Supports & 52.09 \\ 
        CLIP-large & Always Refutes & 49.59 \\ 
        CLIP-large & Always NEI & 49.63 \\ 
        CLIP-large & Guided & 51.64 \\ 
        CLIP-large & Oracle & 61.38 \\ \hline
        CLIP-large-336 & baseline & 50.98 \\ 
        CLIP-large-336 & Random & 50.04 \\ 
        CLIP-large-336 & Opposite & 53.07 \\ 
        CLIP-large-336 & All & 45.54 \\ 
        CLIP-large-336 & Always Supports & 53.77 \\ 
        CLIP-large-336 & Always Refutes & 51.23 \\ 
        CLIP-large-336 & Always NEI & 51.64 \\ 
        CLIP-large-336 & Guided & 54.22 \\ 
        CLIP-large-336 & Oracle & 62.61 \\ \hline
    \end{tabular}
    \caption{ The Mocheg results with CLIP models with explanations generated by GPT-4o.  }
\end{table*}

\end{document}